# A Review of Web Infodemic Analysis and Detection Trends across Multi-modalities using Deep Neural Networks

Chahat Raj[1], Priyanka Meel[2]

Department of Information Technology

Delhi Technological University-110042, New Delhi, India

[1]chahatraj58@gmail.com, [2]priyankameel86@gmail.com

**Abstract:** Fake news and misinformation is a matter of concern for people around the globe. Users of the internet and social media sites encounter content with false information much frequently. Fake news detection is one of the most analyzed and prominent areas of research. These detection techniques apply popular machine learning and deep learning algorithms. Previous work in this domain covers fake news detection vastly among text circulating online. Platforms that have extensively been observed and analyzed include news websites and Twitter. Facebook, Reddit, WhatsApp, YouTube, and other social applications are gradually gaining attention in this emerging field. Researchers are analyzing online data based on multiple modalities composed of text, image, video, speech, and other contributing factors. The combination of various modalities has resulted in efficient fake news detection. At present, there is an abundance of surveys consolidating textual fake news detection algorithms. This review primarily deals with multi-modal fake news detection techniques that include images, videos, and their combinations with text. We provide a comprehensive literature survey of eighty articles presenting state-of-the-art detection techniques, thereby identifying research gaps and building a pathway for researchers to further advance this domain.

## 1. Introduction

Since the advent of online social platforms, the need to establish baselines for information-assessment across all information flow channels has demanded researchers' attention. It has been dubious how wisely people using these platforms actively communicate and digest the information circulating on the internet. With the exponential emergence of social media as a globally used platform in the recent decade, we have encountered massive escalation in fake news dispersion.

Any act of deliberate, miscreant, or unverified inclusion of information creates fake news. Substantial instances include widespread misinformation since January 2020 claiming World War 3 [1], its probability, countries involved, and tentative dates. The arrival of the COVID-19 pandemic has led to a multi-fold rise in disseminating fake news globally. This escalated spread of misinformation amidst the pandemic has been termed as 'Infodemic.' Statements such as "Turmeric powder and black pepper cure coronavirus [2]", "Cocaine treats COVID [3]" etc. have appeared as post contents on every social site in the forms of text messages, images, and videos. Such fake news can be found listed in various official websites debunking the claims [4]. Another fake message states a WHO-issued four-step protocol to prevent COVID-19 circulated at large on most online social networking platforms, especially on WhatsApp [5]. Conspiracy theories and tweets conveyed that the stated countries created the novel coronavirus as a weapon to ignite bio-war [6]. Subsequently, fake news about the pandemic became as contagious as the virus itself. A huge source of fake news was the 2016 US presidential elections that have also focused on many fake news researchers. Other such events and examples of fake news are Pizzagate [7], Indian and Brazilian elections [8], Hurricane Sandy [9], spying technology in Rs. 2000 Indian currency notes [10], Citizenship Amendment Act 2019 [11] & Article 370 (Kashmir, India) [12], etc., which have had huge adversarial impacts on the lives of people.

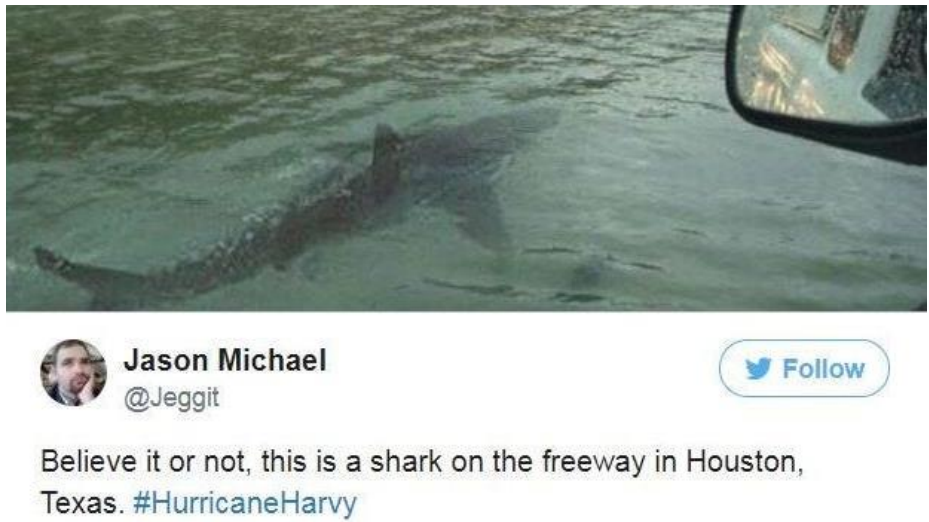

Figure 1: Visual fake news example of a shark in Houston (Source: USAToday)

Social media interactions have contributed a lot towards inventories of big data. Users accord it in text, image, video, audio, emoticons, reactions, etc. Text, images, and videos compose a major actuating portion of the information on the internet and hence blot the grey areas of fake news. Since fake news and its critical nature came into the picture, several scientific developments have been made in textual fake news detection. Fake news detection technologies using NLP, text classification, vector-space models, rhetoric structure theory, opinion mining, sentiment analysis, graph theory, deep neural networks, and others have been created, reviewed, and summarized by fellow researchers [13-16]. Probing of image fake news is comparatively lower, and of videos, it is negligible. Most of the information users encounter on the internet is accompanied by visual representations, either using an image, video, or other modalities. Visual data is quickly gazed upon and leaves a lasting impact. It is the breeding ground of tampering, manipulations, and forgeries. With technology emanating, editing applications and techniques, images and videos are being meddled to mislead information consumers. Moreover, online social media allows users to add to their data, be it real or fake. Recent examples include an image of a shark on a Houston freeway that went viral during Hurricane Harvey (Figure 1). The image was shared and retweeted at large, creating havoc amongst people. Misinformation has been so outrageous that even US President Donald Trump could not escape from it and went on to retweet a fake video that stated anti-malaria drugs could cure coronavirus, which was later brought down when proven fake [17]. In 2020, during the coronavirus pandemic, fake news claiming Kim Jong Un dead, absconding, or assassinated spread widely [18]. Fake videos of his funeral were shared on social media (Figure 2).

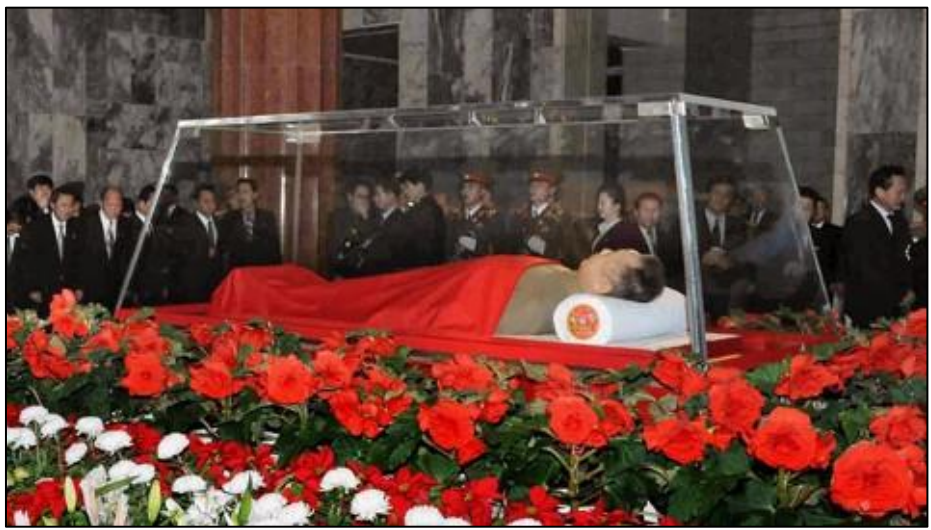

Figure 2: Outdated image of King Jong Un's Death (Source: nypost.com)

In this survey, we broadly categorize data modalities into text, images, and video that are spreading fake news on the internet. We present a literature review in fake news detection techniques that cover these data modalities individually or in combination as hybrid or multi-modal data. This review intends to cover detection mechanisms for data modalities that promote most of the spreading fake news, i.e., text, images, and videos. This survey aims to highlight detection mechanisms that can detect misinformation based on any of these data types. We enlist the challenges and limitations of past research and enlighten ways to future work.

## 1.1 Motivation

As we observe trends among fake news detection, a huge amount of published work and resources are available for text-based detection. Considerable amounts of frameworks and detection mechanisms using textual features have been designed since the problem domain emerged. Machine learning and deep learning algorithms have been applied largely to provide solutions due to their extreme popularity in this domain. These include sentiment analysis, text mining, stance classification, similarity analysis, etc. Texts are analyzed based on their sentence structures, words, punctuations, tone, grammar, and pragmatics. Textual fake news analysis is the domain well explored and worked upon by a large number of researchers. Various other detection techniques have emerged, like using psycholinguistic features, deception modeling, fake news spreading prediction, graph-based propagation detection, and reputation scores.

We present this review in a modality-focused manner. Fake news spreads in the form of textual or visual data. We aim to elaborate on how mechanisms have been developed to deal with each type of fake information that spreads textually through images or videos. The fake news domain has been expressively presented by Sharma and Sharma [19], Figueira and Oliveira [20], Torabi and Taboada [21], Zhang and Ghorbani [22], Tandoc et al. [23], and Shu et al. [24]. Summaries of noteworthy works have been provided in reviews by Mosinzova et al. [25] and Rubin et al. [26]. In contrast, Rajendran et al. [27] brought into the light the importance of Deep Neural Networks for stance classification.

Moving towards fake visual news, tasks performed have been comparatively lower in number. The domain started gaining attention in 2013 and rose considerably from 2017 to 2020. Image fake news detection started gaining importance when image-accompanied news and posts started appearing at large on online platforms. The rise in visual data made ways for fake news to seep in and thus encouraged researchers to explore this domain. Cao et al. [28] have highlighted the role of exploring visual data while detecting fake news. Fake news detection has shown notable performance improvements when a visual analysis is combined with text. Most of the visual classification techniques applied neural networks that provided quick and efficient results. Fake images can be classified as tampered images where some of the other manipulations are made in the image or as misleading images where the context of news content and image do not comply. Some fake news is also accompanied by older images from other events, i.e., combining recent news with outdated images. Another category of fake images that recently emerged is computer-generated images commonly created by Generative Adversarial Networks (GANs). All these types of images contribute to fake news. Any FND framework has not been able to detect fake news that revers to all these types collectively. Individual modules have been created for different tasks. Some architectures can classify tampered images well, while others can spot misleading images that do not match the context. Some frameworks classify images based on statistical features of textual and visual data. Parikh and Atrey [29] provided a survey for multimedia FND. There is a lack of a wholesome framework that could efficiently detect fake news under all modalities. As we talk of visual data, fake news videos have taken up at large over social media. These have a great impact on the minds of viewers and bring adversarial social and political effects. Frameworks for video fake news detection are very few. Multiple features are needed to be studied for video analysis, being a complex task containing spatial and temporal features, speech, and movements. Few researchers have applied it using inconsistencies between speech and lip movements; few have attended analyzing facial expressions following similar trends in real and fake videos, while few utilize the image information at every frame. Other multimedia news in audio, podcasts, and broadcasts are yet less infiltrated with fake news. With an acutely low amount of work performed, video fake news detection is an emerging problem domain that researchers need to pay heed to.

### 1.2 Existing Challenges in Multimodal Fake News Detection

The domain of fake news detection gained popularity very quickly. Large amounts of unverified and uncredible posts have been misleading people. Using linguistic features for credibility assessment of content is a popular and widely-used method. Here, we list the existing challenges to detect fake news spreading through all types of data.

1. **Visual Fake News Detection:** The fake news menace is rising. It began with spreading through text and has now started gripping users through all forms of multimedia. Visual data probable to fake news exploitation can be categorized into images or videos. There are various existing approaches to detect text-based fake news. However, visuals play a great role in impacting viewers' minds and therefore are being infiltrated with fake news in the current generation. An image or video can be easily modified using media editing applications. Various manipulations in visual data go unrecognized through the viewers' eyes. It is very difficult for humans to observe minor changes in modified images and videos to classify them as fake or real. Automated tools are required which can identify minute variations created made to fabricate or manipulate visual data. This poses a great challenge for researchers in designing visual-based fake news detectors.

2. **Auditory Fake News Detection:** Auditory fake news is a type of fake news that has been in existence but is going unnoticed. Many social applications allow users to share recorded audios. These audios files are vulnerable to spreading fake content, propaganda, unverified information, and more. This type of multimedia has not been put into use for credibility assessment. There are no fake news detection mechanisms that incorporate audio as a sole modality or in combination with other data modalities. This issue needs to be addressed to prevent the contamination of audios with false content and serve its early detection.

3. **Detection of Embedded Fake Content:** When one type or modality of data is fixed or embedded within another type of data, it is embedded content. A new type of social media posts is spreading widely known as a 'meme.' It is an image or a video, mostly with text embedded on it. Various forms of media like text, image video, gifs, or hyperlinks are embedded into other forms. It is a complex task to analyze media that is embedded in some other data type. There is an upsurge of fake content in the form of embedded media or memes. Efficient detection mechanisms are required to fight such misleading content.

4. **Multimodal Datasets:** To build fake news detection mechanisms, most machine learning and deep learning tasks require large amounts of data. There is a lack of real-world multimodal datasets. Text-based fake news datasets are more in number than visual or multimodal datasets. Lack of proper datasets limits the extent of research. There is a need to collect real-world fake news data that consists of various types of information like text, image, video, and meta-data.

5. **Holistic Detection Mechanism:** The research community has encountered many techniques that can robustly detect fake news. These techniques use linguistic features, visual features, sentiment scores, social context, network/propagation-based features, meta-data, and hybrid features. There is no such mechanism at present that extracts all these details from a given fake content and predicts its integrity based on all the contributing factors. Different researchers have highlighted the importance of all of these techniques individually or in hybrid combinations. It is worthwhile to consider all these features for building a holistic fake news detector.

6. **Real-Time Verification:** Provided information-spread ease through the internet and online social platforms, fake news is being generated and spread at every instant. Fake news can be about anything and anyone. It spreads continuously as we interact on the internet. Existing detection tools either require users to self-validate a piece of news by fact-checking on their website/application or classify news late after it has been spread and affected various aspects of life. The world needs a system that analyses content in real-time and declares it as fake or real based on its decision.

7. **Lack of Literature:** There is a lack of significant literature in the domain of multi-modal fake news detection. Although many authors have presented textual detection mechanisms, work done in the multimodal domain is minimal and not complete. In this paper, we endeavour to cover all the past research performed using multiple data modalities other than only text-based techniques. We highlight works that have used visual content, alone or along with textual content, for detecting fake news.

## 1.3 Review methodology and data analysis

We provide an enriched and systematic literature review of all the relevant and available articles in the multi-modal and visual fake news detection domain. The articles have been extracted from reputed digital libraries that include Google Scholar, IEEE Xplore, Elsevier, Springer, ACM, and Wiley.

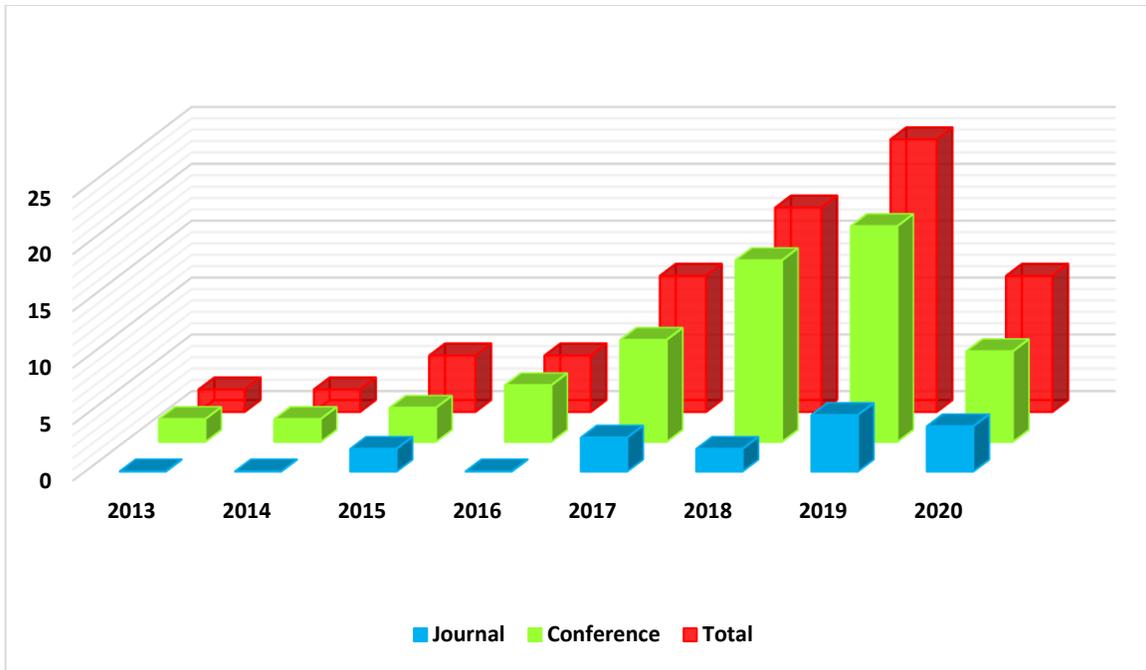

Figure 3: Year-wise trend of published work

We studied over 200 articles on the topic from the libraries mentioned above and shortlisted 70 relevant articles suitable for this survey that display state-of-the-art performance. We have covered all important articles and literature published in various journals, conferences, book chapters, and student thesis reports. Figure 3 provides an overview of the trend of works published yearly between 2013 and 2020. All the related research has been performed, and relevant articles have been published in this span. Some of the brilliant works done have been published in reputed journals or conferences. From the figure, we can observe that focuses moved towards visual and multi-modal fake news detection largely after 2016. The number of published works has been rising since then. It is easily recognizable that tasks involving visual data in the fake news domain has been rising for the past five years and is grabbing researchers' attention.

The search words applied for querying digital databases include multi-modal fake news detection, image fake news detection, fake news detection, multi-modal fake news datasets, and their synonyms. Figure 4, presents the article-wise percentage distribution of algorithms and techniques utilized for multi-modal fake news detection.

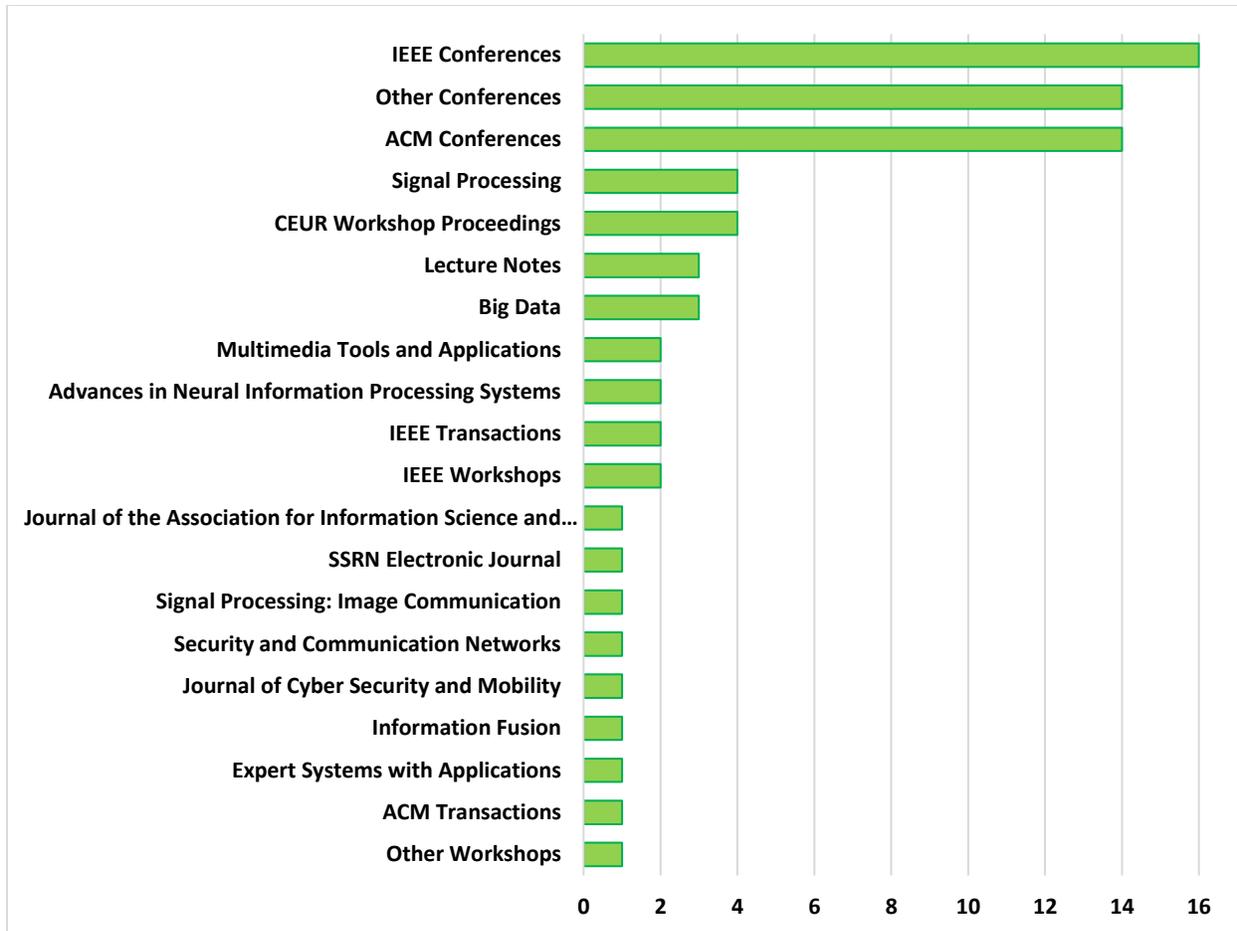

Figure 4: Number of articles published in journals, conferences, and lecture notes

## 1.4 Contribution and Organization of this paper

To the best of our knowledge, this is the first data modality-based review in fake news detection. The contributions of this study are as follows:

- Analyzing and identifying the techniques utilized in multi-modal fake news detection tasks.
- Comparing these techniques based on their applications, advantages, and disadvantages.
- Providing a comprehensive review of remarkable work done in the domain, discussing popular techniques, datasets used, and results obtained.
- Providing a detailed summary of multi-modal usable datasets for fake news detection.
- Comparing the efficiencies of available literature and their work in terms of evaluation parameters utilized.
- Identifying the research gaps in multimodal fake news detection methods and enlisting potential research directions.

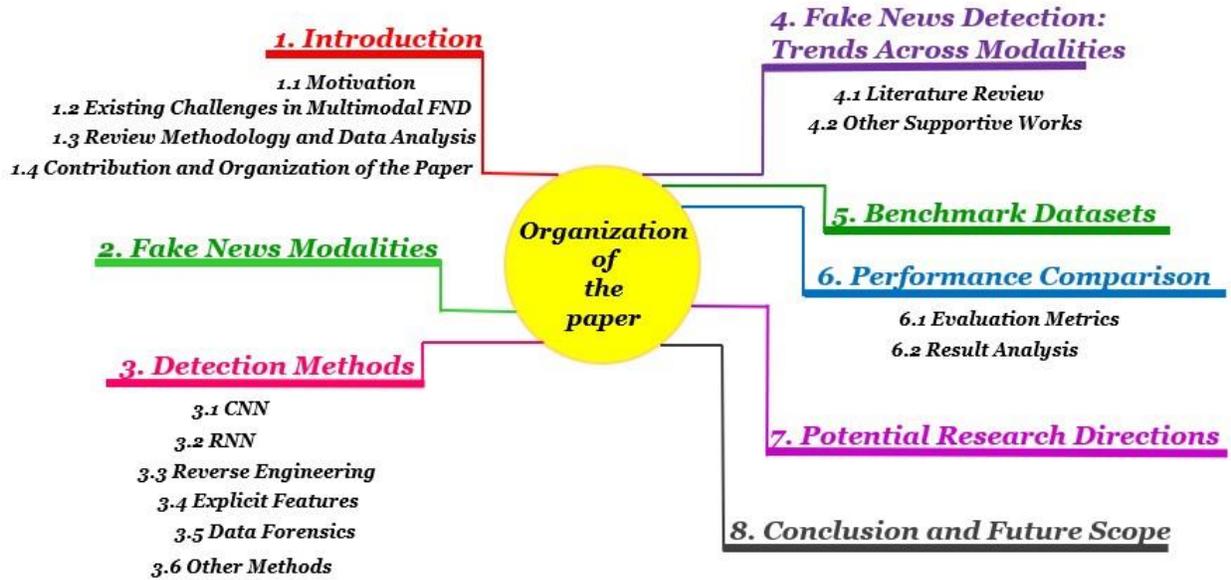

Figure 5: Organization of the paper

The organization of this paper is depicted in figure 5. Section 1 discusses the current Infodemic situation and introduces readers to the problem domain. We describe the motivation to conduct this review and the existing challenges in multimodal fake news detection. The review methodology used and analysis of articles is described next. In section 2, we describe the types of modalities in which fake news propagates, which forms the basis of this review. Section 3 discusses several detection methods used so far across textual and visual modalities and list their advantages, disadvantages, and applications. Section 4 provides a detailed review of the literature and important architectures built for fake news detection tasks. Section 5 serves the readers with rich tabular information of benchmark multi-modal datasets for FND tasks. Section 6 explains the performance evaluation metrics various researchers have used in their works, their distribution, and performance analysis of noteworthy architectures. In section 7, we discuss potential future research directions. Section 8 concludes this review by summarizing our work and imparting motivation and future research directions to readers.

## 2. Fake News Modalities

Fake news is defined as any piece of false information that misleads people. It can be deliberate, fabricated, or simply unintentional. The intent of spreading false news could be maliciously intentional, political, for gaining monetary benefits, popularity, or simply for fun. While referring to data modalities, fake news spreads through text, images, videos, audios, hyperlinks, embedded content, and hybrids. Because of less or no work in the remaining modalities, research is limited to textual and visual modalities. Therefore, we consider these modalities for review, which have been explored by researchers for fake news detection.

**Text:** This is the most popular mode of communication on the internet. People interact through textual matter on social media platforms, websites, blogs, e-mails, personal messaging, and more. Most of the false information spreads through text on the internet. Fake news is found propagating on social media posts, articles, and online messaging services. Text is the simplest and most used way for an internet user to convey his concerns. Being a largely used modality for communication, it also accounts for a large amount of fake news. Figure 6 shows an example of fake textual news. The screengrab is taken from Twitter. In the tweet, the user falsely attributes a claim to WION News, which states that China is hiding the real numbers of death amidst the coronavirus pandemic. The post says that SO2 concentration around Wuhan, China, has grown due to the burning of a large number of dead bodies. The claim is false and has been debunked by various fact-checking websites.

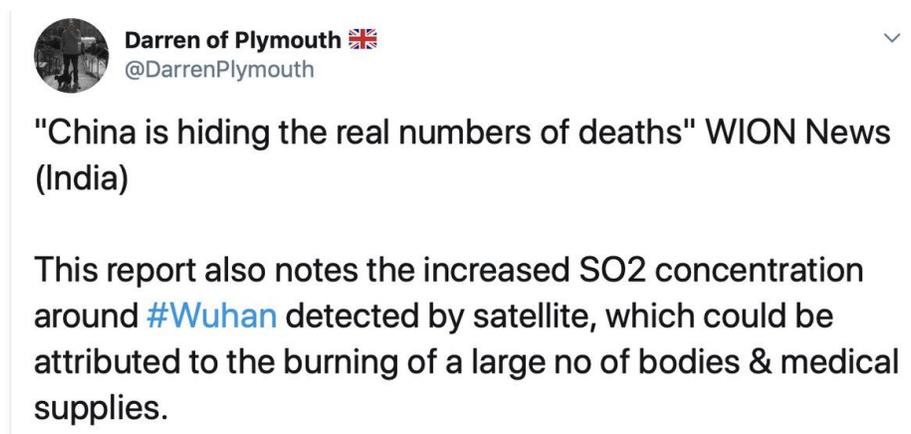

Figure 6: Textual fake news example (Source: Twitter)

**Image:** An image is a visual representation of something. Images are highly vivacious and impactful for depicting anything. They leave lasting impressions on the minds of viewers, whereas words can be forgotten shortly. They have suddenly gained popularity with the increase in the feasibility of sharing them. Images go through certain manipulations to carry a false message. The use of photo editing tools supports these manipulations. Some examples of editing techniques are cropping, splicing, copy-move, retouching, or blurring. Any image can be manipulated to convey a false message, which contributes to fake news. Often, they are not manipulated but accompanied by false text. Many of the times, irrelevant or out-of-context image is placed with fake text. All of these types of images emanate false conceptions accord with fake news. In figure 7, a Facebook post shows a girl rescuing a koala bear from Australia's bushfires. Originally, the picture is a digitally created artwork used out-of-context to match the bushfire situation.

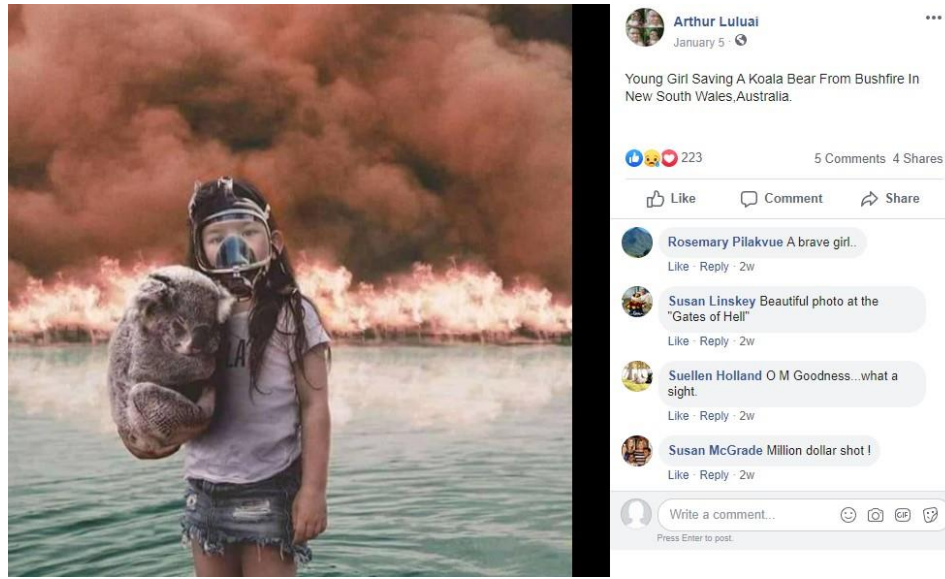

Figure 7: Example of fake news image (Source: Facebook)

**Video:** Sharing of videos online became immensely easy with the introduction of YouTube. The platform allows the feasible sharing of information through video content. Social media platforms allow various techniques for communication through videos. This can be done in the form of regular posts, stories, ads, or even comments. This viability of video interaction gives room for sharing of fake content through videos. Videos are a powerful and impactful tool. They are capable of successfully manipulating people through their content. Therefore, it raises a serious concern to authenticate video content and decide whether a video is credible or not. Figure 8 shows a screengrab from Twitter that shows a video claiming Vladimir Putin's daughter was getting the first shot of coronavirus vaccine. The girl is, in fact, a volunteer and not the Russian President's daughter.

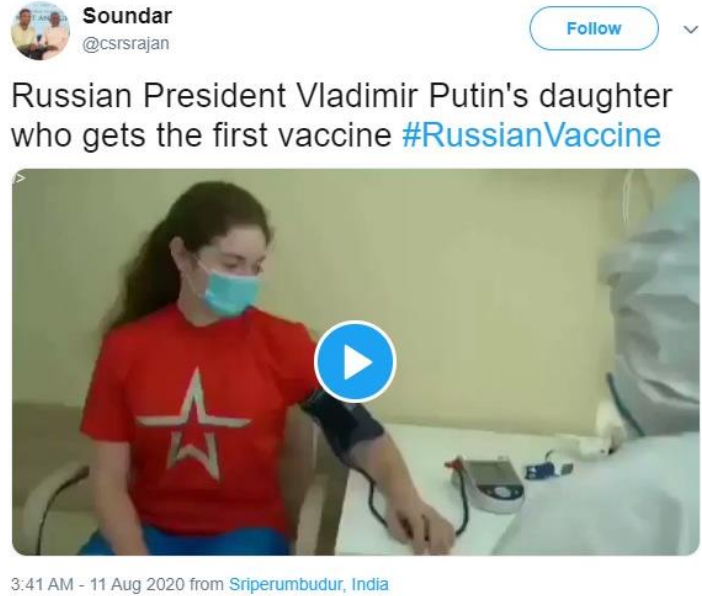

Figure 8: Example of fake news video (Source: Twitter)

**Other Modalities:** Numerous data types have not been analyzed for fake news yet due to limitations of exposure and datasets. These include embedded data types, audios, and hyperlinks (clickbait). Embedded media is that where one type of data is merged or superimposed onto another. For example, textual matter on images or videos, embedding audio in images, altering audio in videos, etc. Detection of fraudulent content in such a data type is complex and challenging. There is a lack of past research in this area. Figure 9 shows a meme with a text embedded on it that says that the North Korean leader Kim Jong Un faked his death to expose traitors. Many such false statements and claims circled the internet. Various fact-checking sites have debunked these claims.

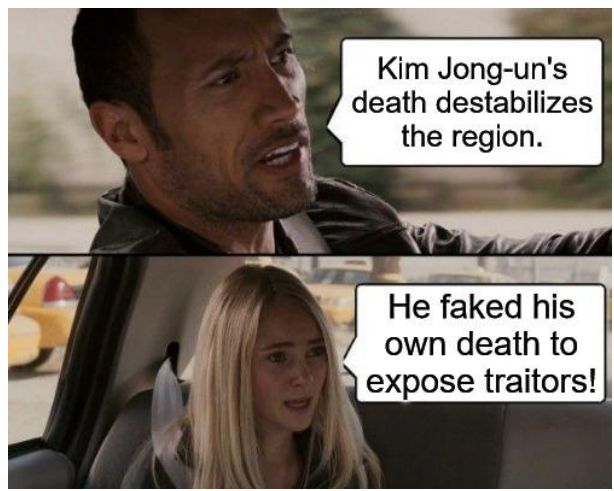

Figure 9: Fake example of a text-embedded image (Source: Facebook)

## 3. Detection Mechanisms

Various fake news detection algorithms utilized by researchers are explained below. Figure 10 depicts the percentage distribution of used algorithms and methods in the reviewed articles. These include Deep Neural Networks, Image/Video Forensics, explicit features, and other methods. Textual detection mechanisms have employed several machine learning and deep learning classifiers. They have been successful enough in this domain. For visual analysis, the most popular tools are Deep Neural Networks. Many researchers have combined these with the use of explicit features available on the web. These can be statistical features, user-based features, post-based features, propagation features, or more. Another popular method is the use of reverse image search on search engines. This method is useful in identifying the integrity of a particular image. Tools are available that use this mechanism to allow the verification of fake news. Several other methods of image and video manipulation have started being merged with fake news detection. This has bridged the gap between methodologies and brought the problem domain and possible solutions in a common range. The applications, advantages, and disadvantages of the following methods have been summarized in table 1.

### 3.1 CNN

First introduced in the 1980s, CNNs have come a long way in the domain of computer vision. They have been applied to Natural Language Processing [30], image classification [31], video classification [32], object recognition [33], time-series forecasting [34], anomaly detection [35], speech analysis [36], handwriting recognition [37] and the likes. For image classification, CNNs require training over large image datasets. Their learning process occurred to be substantially faster than previous methods known, an underlying feature that brought CNNs into the picture. They are efficient in analyzing latent features present inside an image or a video. A rich survey of the latest Convolutional Neural Networks has been provided in Khan et al. [38]. As compared to images, little work has been done in video classification using CNNs as videos are more complex to process owing to their temporal dimension. Most works utilize CNNs to classify videos by extracting images at every video frame [39]. Another method treats spatial and temporal domain separately and classify them using two convolutional neural networks and fusing them after that. Many researchers have also applied CNNs for text classification using one-dimensional convolutional networks.

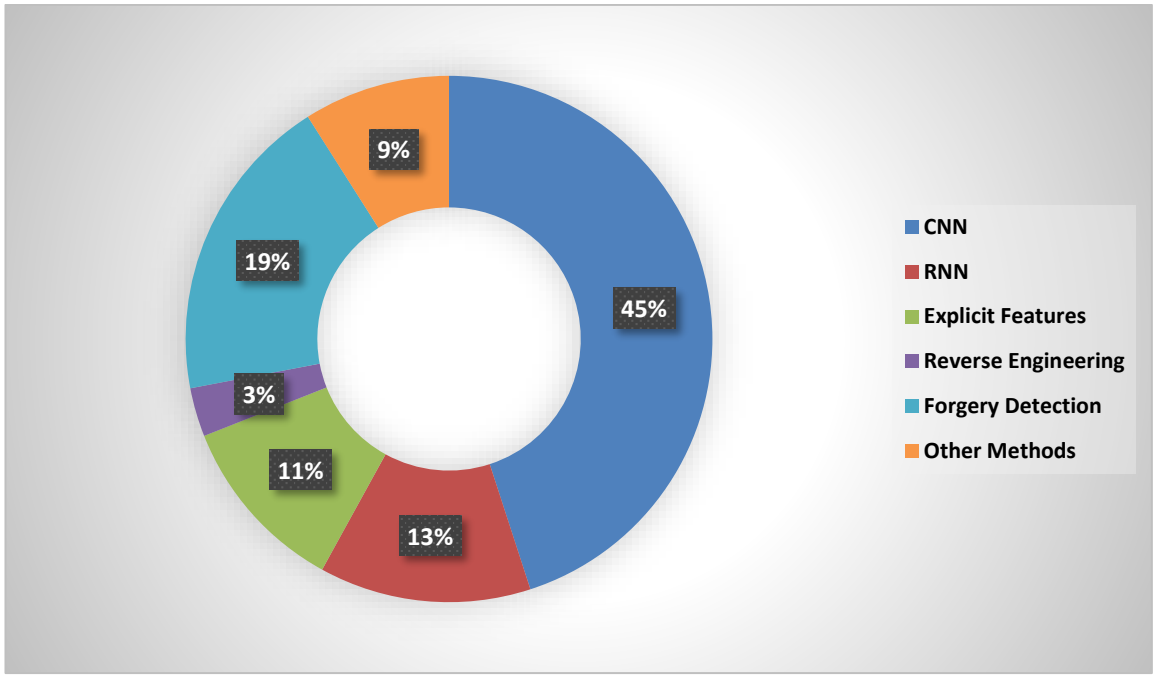

Figure 10: Distribution of reviewed articles representing each method

## 3.2 RNN

Recurrent Neural Networks analyze sequential inputs like text, image, speech, video, and output in a feedback loop. A network with feedback loops is created, which allows RNNs to retain information and train themselves. RNNs have been utilized in image classification [40], video classification [41], object recognition [42], video annotation [43], time-series prediction [44], anomaly detection [45], sentiment analysis [46], speech recognition and other ML and DL tasks. LSTMs (Long Short-Term Memory networks), a special case of RNNs, have been widely utilized in fake news classification tasks in multiple modalities. These train much faster and can perform complex classification tasks than other RNNs. 13% of the articles reviewed in this paper performing multi-modal fake news detection have utilized RNNs and a combination of RNNs and CNNs.

## 3.3 Reverse Engineering

A popular method for fake news detection is to look up the content in question on the internet. Post's credibility is assured by matching it with the occurrences that appear in the search results. The methodology used under reverse engineering for fake news detection is Reverse Image Search. Search engines like Google, Yahoo, and Bing allow their users to input a query image and provide them with relevant information about the image. This process is utilized in fact-checking the authenticity of an image. We can get to know how old an image dates back and where did it appear first. Metadata can also be extracted from such visual data. It also helps us verify the context of the image to the text it accompanies. This method is used by automated fake news detection tools, applications, or web plugins.

Table 1: Comparative analysis of fake news detection methods

| Domain | Method | Applications | Advantages | Disadvantages |
|---|---|---|---|---|
| Deep Learning | Convolutional Neural Networks | Image, video, object, speech, time-series, handwritten data recognition, text mining [30-37] | Adaptively learn classification features | Cannot detect image manipulation on its own Require large labelled datasets |
| Deep Learning | Recurrent Neural Networks | Time, sequence-based predictions, Text, speech, image, video processing, object recognition [40-46] | Great memorizing capacity, usable with CNNs | Slow computation, difficult training, hard to process long sequences, exploding and vanishing gradient |
| Reverse Engineering | Reverse Image Search | Image origin tracking, metadata tracking | Allows verifying variations in images by matching with originally occurring images and text | Can work with images previously present, cannot detect every type of fake image |
| Explicit Feature Analysis | Statistical Analysis, Semantic score analysis, User-profile feature analysis, Propagation feature analysis, Geolocation, Psycholinguistics | Explicit feature detection in all modalities based on features present out of the content | Provides prominent non-data features, measures semantics and relevance | Does not include latent features |
| Data Forensics | Image/video tampering detection, Face manipulation detection | Forged image and video detection, face-swap detection, deepfake detection, editing detection | Easily detects changes made in an image area by editing, removal, addition etc. | Difficult to detect minute manipulations |

### 3.4 Explicit Features

Under explicit features utilized for FND tasks, we categorize statistical features (no. of words, likes, shares, retweets, comments, reactions, etc.), similarity features that analyze the similarities between content and visual information of a news article and state how well both of these are correlated, semantic features that verify meaningfulness of data, user profile features that provide information about users' age groups, backgrounds, faiths and beliefs, inclination, online social behavior, and other relevant profile information, propagation features that help analyze the flow of fake news among networks and people, geolocation features those study areas of fake news generation and propagation and other external features. These features, when combined with other modalities, increase the weightage of detection accuracies. They serve as an important factor for fake news analysis and detection.

### 3.5 Data Forensics

Images and videos, given the current technological advancements, can be easily edited and tampered with. We have classified fake news detection techniques using forgery detection, splice-detection, copy-move detection, face-swapping detection, face manipulation detection, pixel-based forgery detection, photoshop detection, object-removal detection, repurposing detection, and other similar editing detections under image forensics. This method verifies the credibility of images and videos without a requirement of their original version. Algorithms utilized can detect manipulated regions in images and videos. Popular literature that uses data forensics techniques for image and video classification is explained in section 4.2.

### 3.6 Other Methods

Few other methods that have been utilized by researchers for fake news detection can be named as co-occurrence matrices, blockchain, pattern recognition, etc. It has become popular to match the semantics between post text, image, and video. Few of the latest works verify if the post's modalities convey the same meaning and then classify them as real or fake. They have provided a new dimension to investigate fake news detection. This area provides opportunities to be explored more, enhance currently available methods, and leverage new ones.

## 4. Fake News Detection: Trends across Modalities

This section presents an overview of crucial research performed in visual and multi-modal fake news detection. We highlight the vast usage of Deep Neural Networks and forgery detection techniques in multi-modal analysis through the survey. We present the survey classified based on the modalities used for fake news detection.

Qureshi and Deriche [47] explained the taxonomy of types of forgeries found in images: copy-move forgery, image retouching, resampling, image splicing. They have also discussed pixel-based forgery detection methods in images that include contrast enhancement detection, sharpening filtering detection, median filtering detection, resampling detection, post-processing editing detection, copy-move detection, and image splicing detection. Brezeale and Cook [48] provided a survey of existing video classification methods that classify videos using text features, audio features, visual features, and combinations. Boididou et al. [49] have reviewed various methodologies for classifying multimedia data on Twitter that include verifying cues, assessing the source and user credibility, content credibility, image forensics, verifying multimedia use task and have described the verification approaches used, namely UoS-ITI, MCG-ICT, and CERTH-UNITN. Anoop et al. [50] deeply studied fake news detection methods on textual modality, image modality, network modality, temporal modality, and knowledge-based approaches. They have also discussed popular datasets for use. Tolosana et al. [51] reviewed existing face-manipulation detection methodologies. Saini et al. [52] neatly summarized supervised, semi-supervised, and unsupervised multi-modal FND frameworks that include baselines like MVAE and EANN. They have compared state-of-the-art by nicely tabulated data.

### 4.1 Literature Review

Image tampering has become easier than ever, given the advances in photo editing tools. It is crucial to detect such forgeries to keep a check on fake news data. Fake images accompanied by fake news are categorized as in figure 11. The categories are: tampered/edited images, outdated images used with a later situation, and images out-of-context with the accompanying text. Figures 12, 13, 14, and 15 show examples of fake news images. Approaches utilized to detect these fakes include supervised deep learning techniques that require huge training data. Deep neural networks have been successful in the classification of manipulated images from the originals. Table 2 summarizes all the tasks related to fake news detection that involve visual modalities. It helps to understand the necessary details of the related works easily. In table 3, a summary of supportive works is provided, which utilize data forensics mechanisms to identify tampered visual data.

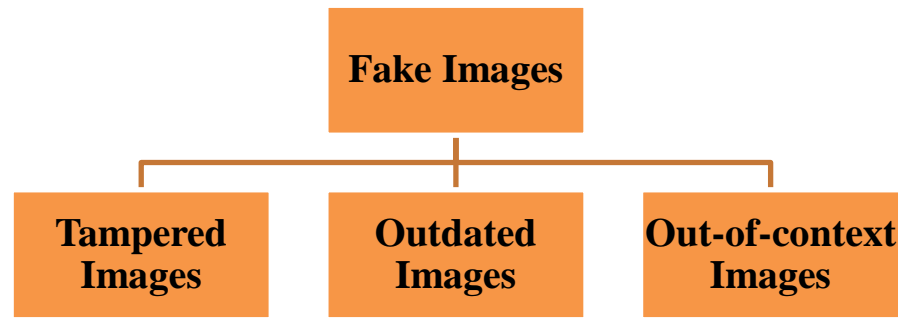

Figure 11: Types of fake images

Elkasrawi et al. [53] allowed users to verify images present online using metadata and feature analysis automatically. The approach has two phases: first involves checking the image for its authenticity, whereas, in second, it is verified if the image has any alterations. The first phase attempts to find other occurrences of an image on the web using Google Image Search. This results in the appearance of the query image in its similar or manipulated versions. Along with several versions of the image, metadata like URL of the image, article in which it appears, publishing date, time, thumbnail, etc., are retrieved. The image's authenticity is validated by matching the timestamps of the query image and search result image. If the image dates back from the date of the query image, it is regarded as fake. In several occurrences, the algorithm uses k-means clustering for the resulting images and their timestamps. In the second phase, image matches are retrieved, and alteration detection is performed. This is done using edge comparison and checking image alignment. The algorithmic features are conglomerated and deployed in a Chrome browser extension for image-based fake news verification.

Wang et al. [54] built a neural network framework for incoming real-time events. This Event Adversarial Neural Network (EANN) framework can handle event-invariant features, thus allowing the detection of fake news on freshly arriving events. With three components in the framework, the first component is a multi-modal feature extractor. Text features are extracted using TextCNN, and a pre-trained VGG-19 is used for visual feature extraction after fine-tuning the hyperparameters. The second component is the fake news classifier, which is built upon the multi-modal feature extraction layers. Classification is performed using a softmax layer for predictions. The third component is an event discriminator that classifies a post into one of the K events. Its application is to remove event-specific features from the posts and capture only invariant features across all events. EANN is capable of classifying fake news incoming from any type of event.

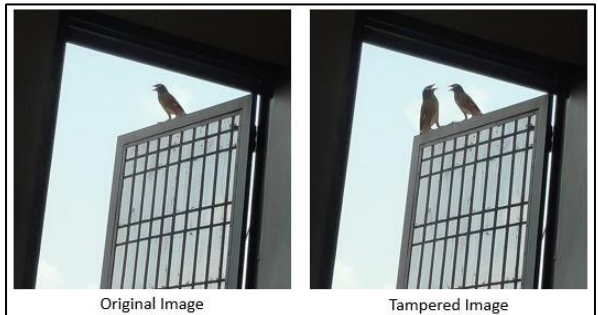 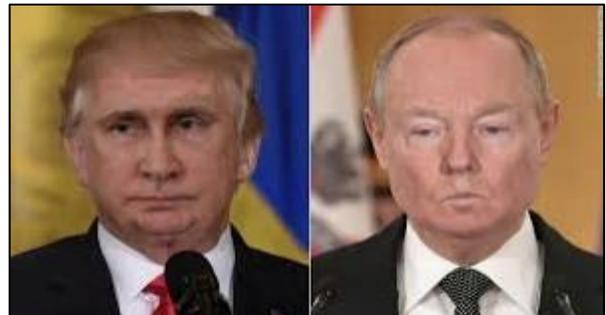

Figure 12: Original vs. tampered image                Figure 13: Face-swapped images

All other image classification tasks that use deep neural networks utilize CNN models, which have been pre-trained on the ImageNet dataset. Jin et al. [55] highlight that to use CNNs for fake news classification tasks, it is required to train the models on a specific fake news dataset. For image credibility analysis of online news on visual data, they collect an auxiliary dataset of fake images from tweets with 328 K rumor images. These images are labeled in terms of their credibility polarity. A collection of 0.6 M images is used to pre-train a convolutional network with architecture similar to AlexNet. The model is trained using iterative transfer learning. This domain transferred learning algorithm provides favorable results for fake news classification.

Qi et al. [56] are the first to use multi-domain information to classify fake and real images proposing MVNN (Multi-domain Visual Neural Network). They expressed that fake news images are constructed of different physical and semantic features than in real news images. They fused the frequency domain and pixel domain feature extraction sub-networks using an attention mechanism to identify real and fake images. Their proposed network consists of three components: a CNN-based frequency domain sub-network, a pixel domain sub-network built with CNN-RNN to extract semantic features, and a fusion subnetwork. As a whole, their network classifies fake news by using only image features, disregarding linguistics, which is a limitation of the task.

Vishwakarma et al. [57] performed FND by scrapping and authenticating web searches. The work proposes an architecture that enables the verification of text-embedded images. Extraction of text from images is supported by Optical Character Recognition (OCR). The extracted text undergoes Name Entity Recognition steps to obtain named entities mentioned in the text. These strings are used as queries for Google Search to find matching results. The resulting links are categorized into reliable or unreliable. Next, the entities extracted from images are checked against the search result articles' titles, further classifying them into real or fake.

Pasquini et al. [58] also verified the integrity of online news. They have also demonstrated that fake news detection in visual has a dependency on visual forensics. Their focus is to detect news web pages that contain misleading images. Their framework automatically detects related news articles on the internet with similar images of an event. The detection is performed using metadata features present along with the images and by comparing similar features in the set of related images.

Using image forensics techniques for fake news classification, Huh et al. [59] utilized self-learned consistency to detect image manipulation and splicing. It is done by comparing patches of images. Tampered images show low consistency scores between patches. Models are evaluated for splice detection and splice localization on unannotated images. It is an efficient way to detect fake/manipulated images where copy-move forgery, object addition, or removal have been done.

SAME (Sentiment-Aware Multi-modal Embedding) is a deep end-to-end embedding network that exploits users' sentiments to classify fake news. Cui et al. [60] have incorporated users' sentiments with images, profiles, content, and comments into a multi-modal framework. The model intakes post content, image, and user profile and feeds them to a Multi-Layer Perceptron (MLP), pre-trained VGG-19 network, and another MLP, respectively. The three modalities are fused using an adversarial mechanism.

Jin et al. [61] performed a classification between real and fake tweets using a two-level classification model: topic level and message level. They proposed that tweets among themselves have a strong relationship in terms of event or topic, and tweets clustered in a topic have similar credibility values. It provided a better result than uni-level classification. Tweets with similar images (verifying by features such as resolution, image popularity, etc.) were clustered under the same topic. They extracted topic-level features and message-level content features, user features, and other available features and classified them on both levels. Topic-level classification

Table 2: Summary of crucial work

| Reference | Task | Modality | Technique | Classification | Dataset | Accuracy /F1 Score |
|-----------|------|----------|-----------|----------------|---------|--------------------|
| [61] | Tweet labeling for FND | Text, Image | Two-level classification (Topic and message-level) | Binary | Twitter (Mediaeval 2015[1]) | 0.94 (F-score) |
| [62] | Fake news detection | Text, Image | Utilizing user-profile features | Binary | FakeNewsNet | >0.90 (F-score) |
| [22] | Fake news detection | Text, Image | BERT, VGG19 | Binary | Twitter, Weibo | 77.77%, 89.23% |
| [55] | Fake news detection | Image | Image splice detection, splice localization | Binary | Columbia, Carvalho, Realistic Tampering, In The Wild, Hays and Efros | 0.91 (mAP) |
| [65] | Fake tweet and its user identification | Text, Image | Reverse image search, User analysis, crowdsourcing | Multi-class (fake, legitimate, not sure) | Twitter | |
| [66] | Fake image classification of Hurricane Sandy | Text, Image | Temporal analysis, Naive Bayes, decision tree | Binary | Twitter | 97% |
| [67] | Image tampering detection, text-image coherence detection | Text, Image | Image forensics | Binary | Mediaeval2016, BuzzFeedNews, CrawlerNews | >75% |
| [68] | Fake news detection | Text, Image | CNN | Binary | TI-CNN | 0.92 (F-score) |
| [110] | Microblogs news verification | Image | Visual and statistical features | Binary | Sina Weibo | 83.6% |
| [56] | Image verification | Image | Metadata, feature analysis | Binary | | 72.7%, 88% |
| [111] | Fake news detection | Text, Image | Bi-LSTM, VGG19 | Binary | Twitter, Weibo | 74.5%, 82.4% |
| [72] | Video annotation | Video | DCT, HSV, SURF, AOF, ResNet, GoogLeNet | Binary | Twitter | 0.40 (F-score) |
| [112] | Video verification | Video | Contextual Cues | Binary | IVC, FVC | 0.9 (F-score) |
| [97] | Fake news detection | Text, Image | VGGNet, sentiment analysis | Binary | PolitiFact, GossipCop | ~75, ~80 (Macro, Micro F1) |
| [92] | Fake news detection | Image | CNN, Iterative Transfer Learning | Binary | Weibo | 77% |
| [113] | Rumour detection on microblogs | Text, Image | LSTM, att-RNN, VGG19 | Binary | Twitter, Weibo | 78%, 68% |
| [93] | Fake news detection | Image | CNN (frequency domain), CNN-RNN (pixel domain), Bi-GRU | Binary | Twitter, Weibo | 84.6% |
| [91] | Fake news detection | Text, Image | Text-CNN, VGG19 | Binary | Twitter, Weibo | 71.5%, 82.7% |
| [114] | Fake news detection | Text, Image, Source | CNN | Binary | Twitter | 82.47% (F-score) |
| [115] | Semantic Integrity Assessment | Text, Image | MAE, Bi-DNN, VSM | Binary | MAIM, Flickr30, MS COCO | 0.75, 0.89, 0.94 (F-scores) |
| [100] | Fake News Detection | Image | Montage detection (feature-based approach), SIFT, SURF | | COCO, INRIA | >90% |
| [116] | Fact-Checking on image-claim pair | Text, Image | Cosine Similarity, Embedding similarity | Binary | Snopes, Reuters | 80.1% |
| [117] | Fake news detection | Text, Image | Machine Learning Algorithms | Multi-label | Kaggle Fake News Dataset 2017 | 85.25% |
| [70] | News consistency verification | Text, Image, Location, Events | CNN | | TamperedNews, News400 | 94% |
| [69] | Fake News Detection | Text, Image | Text-CNN, Visual, Similarity features | Binary | PolitiFact, GossipCop | 87.4%, 83.8% |

Results were fused into message-level feature-vector as an extra feature and then trained the classifier. Each tweet was given a separate label instead of labeling each event.

Shu et al. [62] presented a way to utilize user profile features for fake news detection. They extracted and studied explicit and implicit user profile features, also studying which users were most likely to share real and fake news. They have studied users' geolocations, profile images (using CNN, pre-trained VGG16 model), and political bias. PCA was used for dimensionality reduction of profile features. They compared fake news detection performance using UPF (User Profile Features) to multiple approaches, including RST (Rhetorical Structure Theory), LIWC (Linguistic Inquiry and Word Count), RST_UPF, and LIWC_UPF concluding that UPF and UPF-allied techniques provided higher accuracies than others. They also proved that implicit and explicit features, when combined, provided greater results than each being used individually.

Ajao et al. [63] applied hybrid CNN and LSTM-RNN models for text and image classification. LSTM RNN was used to process and classify text sequences. Another model used was LSTM along with dropout regularization 0.2 to remove over-fitting. The third model incorporated a CNN layer after the word-embedding layer of the LSTM model. The models plain vanilla LSTM, LSTM-CNN model, and LSTM with dropout regularization performed in the order of decreasing accuracies 82%, 80%, and 74%, respectively. Under-fitting and lack of sufficient training data account for the low accuracy of the LSTM-drop model. They also showed that hybrid deep learning models' usage provides considerably good accuracy without requiring huge training data.

Singhal et al. [64] have designed a model named SpotFake for detecting multi-modal fake news without any subtasks like finding correlations between textual and visual data. The model consists of a textual feature extractor module that uses BERT (Bidirectional Encoder Representations from Transformers), a visual feature extractor module that uses VGG19, and a fusion module using simple concatenation. The results outperformed state-of-the-art methods EANN and MVAE and provided higher accuracies equal to 77.77% and 89.23% on Twitter and Weibo, respectively.

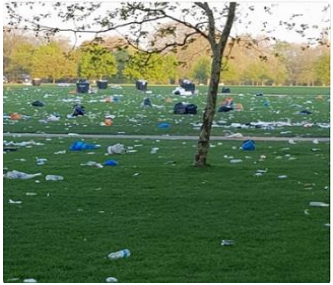

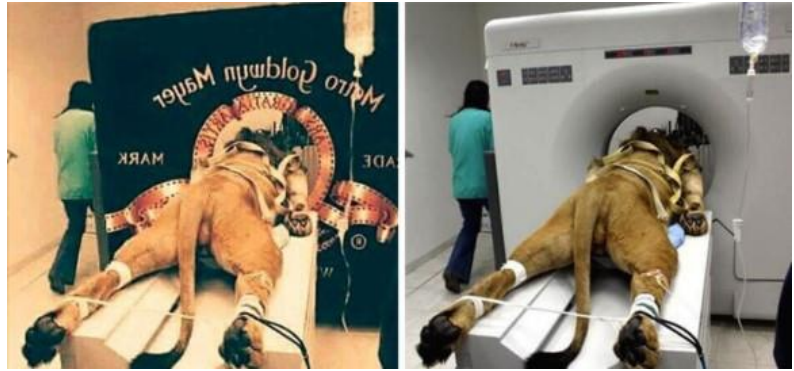

Figure 14: Out-of-context image (Mumbai's park image accompanied with Hyde Park's news), (Credit: theconversation.com)

Figure 15: Out-of-context image (A lion taken to vet displayed as a lion been strapped to capture MGM logo), (Credit: snopes.com)

Trumper [65] developed a web tool, 'Fake Tweet Buster,' for users to check a tweet's credibility. The user needs to enter a tweet URL, and the application provides a result as fake or legitimate. The tool works on reverse image search (Google Images and TinEye), user analysis, and crowdsourcing. The tool provides the user with matching images, image data (old or new), tweet information, and classification result. It allows the tool-user to enter his opinion about the tweet as fake, legitimate, or unsure. This crowdsourced data can be used in the future to provide value to credit score.

Gupta et al. [66] extracted tweet information and images from Twitter related to Hurricane 'Sandy'. They analyzed that retweets contained many fake images (86%) rather than original tweets. The approach used temporal analysis to study the propagation of fake images over the Twitter network considered a graph with multiple nodes. They created graph networks for followers and retweets of each user studied. For classification, they used user-level features (giving poor results) and tweet level features (providing effective classification), experimenting with two machine learning techniques: Naïve Bayes classifier and J48 Decision Tree classifier, providing 91% and 97% accuracy, respectively.

Lago et al. [67] created a fusion of image forensics algorithms to distinguish between real and fake images. The various approaches used are Image Forensics Methods (Error Level Analysis, Block Artifact Grid Detection, Double Quantization Likelihood Map, Median-filter noise residue inconsistencies detection, JPEG Ghosts, Color Filter Array), Splicebuster, and with the use of CNNs. They also verified the coherence of online news with their accompanying images. Texts were analyzed using TF-IDF or its higher version, STF-IDF. To make classifications, they chose Random Forest and Logistic Regression.

Yang et al. [68] used convolutional networks for both textual and visual fake news detection. They analyzed latent features and explicit features of text and image sub-branches and fused them using early fusion to classify news.

Huckle and White [69] utilized blockchain and cryptography to trace the origination of fake content. Their focus is to determine where the fake information comes from rather than analyzing its structure and features. Their approach lies in determining the cryptographic hash functions of the text and associated image. The verification is made by matching the hashes of the versions of an image occurring at different places. Same images from different occurrences resulting in different hashes indicate that the image has been altered. This principle forms the base of their fake news detection mechanism.

Krishnan and Chen [70] identified tweets containing fake news using data mining, statistical analysis, reverse image search, and cross verification. They have divided their framework into two components: Core and Website. The Core module fetches tweets, extracts the feature set, performs classification, and returns predictions. The Website module is mainly responsible for crowdsourcing and collecting users' guesses about the post's credibility. It also returns the final decision about the post to the end-user. Classification is performed using the J48 decision tree classifier and Support Vector Machine (SVM). The crowdsourced data is stored in a database for future re-training of classification models and performance improvement.

The NewsVallum approach introduced by Armano et al. [71] uses text-image semantics for fake news detection. It focuses on multi-modality using text and image features for classification with deep neural networks and reinforcement learning. It evaluates the credibility of news spread online on a daily or hourly basis.

Zhou et al. [72] exploited the similarity between text and images to detect if a news article is true. The proposed model comprises three modules where the first one is a multimodal feature-extractor, the second module is a classifier, and the third module determines the cross-modal similarity between text and image in the post. They have used a text-based one-dimensional convolutional network for text-classification, and image features are extracted using VGG. Both the feature extractors are combined using a concatenation operation. In the third module, the semantic relation between the text and image pair is calculated using cosine-similarity. This helps in identifying if the modalities of a post express the same meaning. A post is classified as fake or real, depending on all of these features.

Chen et al. [73] argue the need for an automatic fake news detector tool for evaluating the integrity of any news online.

Using multi-modal features, Budack et al. [74] measured the consistency between the modalities for fake news verification. The proposed work evaluates the coherence between text and image data in an unsupervised manner. Textual module extracts persons, entities, and locations using Named-Entity Recognition (NER). POS Tagging is applied, and subsequently, embeddings are calculated using fastText. For visual features, the ResNet model is used for verification. Persons, locations, events, and scene context verification is performed in the cross-modal entity verification process. This model applies to real-world news classification.

Parikh et al. [75] performed the task on tweet text and images providing a web application. The UI allows users to upload a screengrab of a tweet from which the model extracts useful information like tweet text, image, username, timestamp, location, etc., and predict the authenticity of a tweet using these features.

It has become easier to create fake scenarios in videos by replacing, removing, or adding objects, adding text, and swapping face in the recent technological eras. These manipulated videos in which a person's face are swapped by another's are termed as 'deep-fakes.' Such videos have created nasty issues of defamation. Forensics based detection methods majorly detect copy-move manipulations or tampered frames. Neural networks have been applied to detect morphed faces in visual data.

Nixon et al. [76] annotated videos as real or fake to verify the news. They analyzed the text of news stories and videos circulating online and used their metadata to fact-check and annotate the videos. In textual analysis, the author checks the stories for correctness, distinctiveness, homogeneity, and completeness and then groups them under clusters. The videos related to these news stories were then retrieved and annotated based upon fragment information.

Bagade et al. [77] developed a fact-checking web and mobile application 'Kauwa-Kaate' for full-article verification incorporating text, images, and videos. Their proposed system provides a user-friendly interface to query and fact-check information as and when they encounter it. The algorithm scrapes news articles from fact-checked and trusted news sources available on the internet and maintains a repository in the backend. The verification is carried out by matching the query item with news articles in the repository. Devoting a platform entirely for fact-checking, is a very practical method for users to verify fake news.

Using several tweet-based and user-based features, Boididou et al. [78] have introduced a model that predicts tweets as fake or real depending upon the majority vote from individual classifiers that use different features. The approach has displayed a successful classification of news from a variety of events.

An event rumor detection mechanism for Sina Weibo has been designed by Sun et al. [79]. The model uses many features, which are content-based linguistic features, user-based features, and multi-media features. The model is suitable for detecting rumors in the form of text, image, and video.

## 4.2 Other Supportive Works

Detecting fake news in visual data is closely related to identifying manipulations in images and videos. The utilization of data manipulation detection algorithms for fake image/video detection offers a worthy scope. In this sub-section, we identify some of the important manipulation detection techniques that could be useful in fake news detection tasks.

A new convolutional layer has been proposed by Bayar and Stamm [80] to classify unaltered and manipulated images. Fake images can be of the misleading type where news content and accompanying image are unrelated or of the tampered type where images are forged to create a fake scenario. Detecting forged images can prove highly beneficial to detect fake news and news containing fake images. The new layer can learn manipulation detection features without needing to extract preliminary features. This layer has been incorporated into a CNN architecture to detect multiple manipulations. For model training, ReLU activation and SGD (Stochastic Gradient Descent) is used. CNNs train faster with ReLU. Binary classification classifies images into unaltered vs. tampered images with 99.31% accuracy. Multi-class classification is used to classify images based on types of forgeries used: median filtering, Gaussian blurring additive white Gaussian noise, re-sampling vs. authentic image with 99.10% accuracy.

Sabir et al. [81] and Jaiswal et al. [82] have detected image repurposing where unaltered images are put together with false metadata in a news item. Pomari et al. [83], Zampoglou et al. [84], and Wu et al. [85] detected fake images by checking if the images or their portions have been sliced.

Table 3: Examples of data manipulation detection techniques

| Reference | Task | Modality | Technique | Classification | Dataset | Accuracy /F1 Score |
|-----------|------|----------|-----------|----------------|---------|--------------------|
| [39] | Deception detection | Text, Video, Audio | 3D-CNN, CNN, openSMILE | Binary | Courtroom trials | 96.14% |
| [105] | Tampered video detection | Video, Audio | Speaker inconsistency detection | Binary | VidTIMIT, AMI, GRID | <1% (EER) |
| [81] | Image repurposing detection | Text, Image, Location | CNN (VGG19), Word2Vec | Binary | MEIR | 0.80 (F-score) |
| [102] | Fake video detection | Video | RNN, CNN (InceptionV3), LSTM | Binary | Deepfake videos, HOHA | >97% |
| [103] | Forged image and video detection | Image, Video | Capsule Forensics, VGG19 | Binary | Deepfake | 99.23% |
| [91] | Face manipulation detection | Video | RNN, CNN (ResNet50, DenseNet) | Binary | Deepfake, Face2Face, FaceSwap | 96.9% |
| [83] | Splice detection | Image | CNN (ResNet50), Illumination Maps | Binary | DSO, DSI, Columbia | 96% |
| [86] | Fake image detection | Image | CNN ensembles | Binary | CelebA, PGGAN | 99.99% (AUROC) |
| [87] | Image translation detection on compressed and uncompressed images | Image | Conventional methods, CNN (DenseNet, InceptionNetV3, XceptionNet) | Binary | CycleGAN | 89% |
| [88] | GANs vs. real image detection | Image | VGG19 | Binary | RAISE, Rahmouni | 100% |
| [99] | GANs Image detection | Image | Co-occurrence matrices, CNN | Binary | cycleGAN, starGAN | 99.45%, 93.42% |
| [80] | Image manipulation detection | Image | New convolutional layer | Binary and multi-class | Various camera images | ~99.10% |

| [104] | Face Spoofing Detection | Image | CNN | Binary | CASIA, REPLAY-ATTACK | <5% (HTER) |
|---|---|---|---|---|---|---|
| [118] | Fake image detection | Image | Mixed Adversarial Generators | | FantasticReality | 0.61 (mAP) |
| [98] | Fake image detection | Image | Color disparities | Binary | CelebA, LFW, generated images | >90% |

Pomari et al. [83] did so by making use of illumination inconsistencies in the image. Computer-generated images and videos have gained huge popularity in the present scenario. It has become fairly easy to generate fabricated content through computers. Such content is entirely unreal or mixed with some kind of real-world entities. The result is a fake product that is not reliable at all. Tariq et al. [86], Marra et al. [87], Nguyen et al. [88], Rahmouni et al. [89], and Rezende et al. [90] have identified fake images generated by computer machines. Nguyen et al. [88] did this using Modular CNN. Sabir et al. [91], Zhang et al. [92], Zhou et al. [93], Dang et al. [94] have detected tampered faces in images, which can be utilized in detecting fake news where faces of celebrities have been swapped to create fake scenarios.

Wu et al. [95] have used supervised learning to trace various types of manipulations like copy-move forgery, object removal, splicing, and other unknown tampering in images. Photoshop is a widely used tool used by content creators to modify visual content. Wang et al. [96] have detected variations created in images using Adobe Photoshop. Wu et al. [97] created BusterNet for copy-move forgery detection. It is a type of forgery where an object from an image is removed from its original location and moved to a different place in the same image. Li et al. [98] used RGB color components in the images to detect changes that occurred due to tampering. Nataraj et al. [99] detected GAN-generated images using co-occurrence matrices. These matrices describe the distribution of co-occurring pixel values or colors. Steinebach et al. [100] recognized image montages for fake news detection. In image montaging, two or more images or their parts are arranged together by cutting, overlapping, pasting, etc., to make a composite image. Korshunov and Marcel [101] and Guera and Delp [102] addressed facial manipulations in videos where the face of a person is replaced by another. They used deep neural networks for the task. Guera and Delp [102] also contributed with a dataset of 300 deep-fake videos extracted from websites. They classified videos using CNN and LSTM into pristine and deep-fake categories.

CNN has been used for feature extraction from video frames, concatenated and propagated to LSTM for analyzing sequences temporally. This architecture allows detecting fake videos as short as 2 seconds of length. Korshunov and Marcel [101] showed that using static frame features correspond to higher accuracies than using audio-visual analysis.

Nguyen et al. [103] identified forgeries like replay attacks, computer-generated images/videos by building a capsule network with CNN layers. Videos are analyzed at frame level, and the probabilities of fake and real of every frame are averaged to generate results for the video.

Yang et al. [104] detected swapped faces in images and videos using CNN. Korshunov and Marcel [105] performed tampered video detection using inconsistencies between audio and video features. Classifiers used were GMM (Gaussian Mixture Model), SVM, MLP, and LSTM. Krishnamurthy et al. [39] performed deception detection over a small dataset of 121 courtroom videos. They used Text CNN for textual analysis, 3D-CNN for videos, and openSMILE with MLP for audio analysis. They also utilized micro-expression features such as smile, laughter, frown as another modality for deception detection. Data fusion techniques used were Concatenation and Hadamard + Concatenation. Wu et al. [106] and Rosas et al. [107] have also proposed deception detection on videos using real-life trial data.

Li et al. [108] detected tampered faces in videos to detect swapped faces of celebrities by detecting eye-blinking features. Eye blinking is detected using the LRCN (Long-term Recurrent Convolutional Networks) model, measuring how much open an eye is measured concerning frame coordinates. Features have been extracted using the VGG16 convolutional layer and propagated to a sequence learning module that uses LSTM-RNN, which can retain memory. Then, the probability is determined of how much the eye is open or closed. Bestagini et al. [109] detected local tampering in video sequences. This was done by finding duplicated frames in videos and cross-correlating them with Spatio-temporal frame regions.

## 5. Benchmark Datasets

Lack of suitable multi-modal datasets have, a lot, hampered the progress in the direction of fake news detection. Deep learning algorithms largely depend on huge amounts of training data, which, being meager, has appeared as a big challenge. The maximum number of fake news detection frameworks built to date have been trained upon data extracted from Twitter, Sina Weibo, or some websites. A few of the other small-sized datasets have been generated for image analysis, which still is limited in number and not of optimum quality. Video datasets for this task are very rare and contain videos in the count of 100-200. Other video datasets are not completely relevant. There is an urgent demand for good quality multi-modal datasets that would furnish the need of the hour. The advancements in data augmentation or computer-generated data are beginning to contribute towards building datasets. For the time being, we present a piece of tabulated information about available datasets (image, video, and multi-modal) that have been used in the above-reviewed articles for fake news detection and similar tasks (Table 4). We also list out such datasets that contain news article URL or image/video URL. These datasets can be further improved by extracting visual data using web scraping methods.

Table 4: Benchmark Multi-modal datasets

| Dataset | Year | Type | Description | Class | Size | Source |
|---------|------|------|-------------|-------|------|--------|
| MediaEval[1] | 2015 | Tweet, Image | Tweets related to 11 events (dev set) and 17 events (test set) | Binary | 6,225 real tweets 9,596 fake tweets | Topsy, Twitter API |
| FakeNewsNet[2] | 2018 | Text, Image | News Content, Social Context, Dynamic Information, article URLs | Binary | 6,000 Fake and 18,000 Real | PolitiFact, GossipCop |
| Fakeddit[3] [119] | 2019 | Text, Image | Text, Image, Metadata and Comment | Multi-class | 8 Lakh samples | Reddit |
| Newsbag [120] | 2020 | Text, Image | News articles and images | Binary | 15,000 fake and 2 lakhs real | The Onion, The Wall Street Journal |
| Newsbag++ [120] | 2020 | Text, Image | Created by Data Augmentation | Binary | 3,89,000 Fake and 2,00,000 Real | The Onion, The Wall Street Journal |
| FVC[4] [112] | 2017 | Text, Video | Fake Video Corpus for fake video detection | Binary | 55 Fake and 49 Real videos | YouTube |
| Twitter [113] | | Text, Image | Text, Image, Social Context | Binary | 6,026 Real and 7,898 Fake | Twitter |
| Sina Weibo [113] | | Text, Image | Data extracted from Chinese online social platform | Binary | 4,749 Fake and 4,779 Real articles | Sina Weibo |
| Politifact [60] | | Text, Image | News Content and their corresponding images, Retweet Comments | Binary | 432 Fake and 624 Real News | Twitter |
| GossipCop [60] | | Text, Image | News Content and their corresponding images, Retweet Comments | Binary | 5,323 fake 16,817 real | Magazines, Newspapers and Social Media |
| CrawlerNews [67] | 2017 | Text, Image | News articles and images | Binary | 2,500 Images | Google News |
| Mediaeval 2016[5] [112] | 2016 | Text, Image | Tweets and images from 53 past events | Binary | 17,857 Tweets with 10,628 fake and 7,229 Real | Twitter API |
| MFN [64] | 2018 | Text, Image | Event Centric dataset of tweets and corresponding images | Binary | 14,000 Tweets and 500 Images 1,154 Real and 1,154 Fake News Articles | Twitter, Snopes and Webhose |
| TI − CNN[6] [68] | 2019 | Text, Image | Text, metadata and Image URLs | Binary | 20,015 total articles with 8,074 Real and 11,941 Fake | Over 240 Websites |
| [117] | 2019 | Text, Image | Metadata, News Articles | Binary | 3,568 Fake, 15,915 Real | Multiple Websites |
| ReCOVery Dataset [121] | 2020 | Text, Image, Social Information | News Articles, Tweets related to CovID-19. | Binary | 2,029 News Articles, 1,40,820 Tweets | Twitter, Multiple Websites |
| TamperedNews [74] | 2020 | Text, Image | News articles and images | Binary | 72,561 News Articles | BreakingNews Dataset |
| News400 Dataset [74] | | Text, Image | Tweets, Articles | Binary | 400 | Twitter, Websites |
| WhatsApp Dataset [8] | 2020 | Image | Fake images extracted from WhatsApp groups | | 8,44,000 | WhatsApp |

---

# 6.  Performance Comparison

In this section, we demonstrate the usage of evaluation metrics utilized by fake news detection tasks and compare their performances based on the most utilized metrics, i.e., accuracy and F1-score. The comparison provided here is irrespective of the dataset but highlights each task's features and methods. We determine how the results have been moving all these years and identify prospective detection methods. Performances are displayed for tasks displaying the results achieved by the experiments on datasets they have used. Visual representations are provided for an easy understanding of how a given model performs when they use a specific set of features.

## 6.1 Evaluation Metrics

This section discusses the various evaluation parameters utilized for fake news classification tasks. We explain the evaluation methods utilized to examine a model's performance using Accuracy, Precision, Recall, F-score, ROC, AUROC, and EER. In figure 16, a confusion matrix is presented that explains the categorization of rightly and wrongly classified items. We refer to the items as news, images, and videos.

| CONFUSION MATRIX | Actually Real | Actually Fake |
|---|---|---|
| Predicted Real | TRUE POSITIVE | FALSE POSITIVE |
| Predicted Fake | FALSE NEGATIVE | TRUE NEGATIVE |

Figure 16: Confusion Matrix

True Positive (TP): This includes rightly predicted positive values, i.e., a piece of real news, image or video is classified as real.

True Negative (TN): This includes rightly predicted negative values, i.e., an originally fake item is correctly predicted as fake.

False Positive (FP): This class contains values where a fake item is wrongly predicted as real.

False Negative (FN): This contains real items wrongly predicted as fake by the model.

$$Accuracy = \frac{TP+TN}{TP+FP+FN+TN}$$

$$F1\ Score = \frac{2(Recall \times Precision)}{(Recall+Precision)}$$

$$Precision = \frac{TP}{TP+FP}$$

$$True\ Positive\ Rate\ (TPR) = \frac{TP}{TP+FN}$$

$$False\ Positive\ Rate\ (FPR) = \frac{FP}{TN+FP}$$

$$Recall = \frac{TP}{TP+FN}$$

Table 5: Evaluation metrics used in reviewed articles

| References | Accuracy | Precision | F1 | Recall | AUC | EER | Others |
|---|---|---|---|---|---|---|---|
| [85], [61], [67], [68], [70], [76], [81] | | ✓ | ✓ | ✓ | | | |
| [54], [55], [56], [97], [62], [63], [67], [72], [75], [112], [113], [110], [111] | ✓ | ✓ | ✓ | ✓ | | | |
| [80], [53], [102], [116], [96] | ✓ | | | | | | |
| [101] | | | | | | ✓ | |
| [59], [84], [60] | | | | | | | ✓ |
| [115] | | | ✓ | | | | |
| [93], [94], [108], [109], [106] | | | | | ✓ | | |
| [89], [74] | ✓ | | | | ✓ | | |
| [49] | | ✓ | ✓ | | | | |
| [92], [117] | ✓ | ✓ | ✓ | ✓ | ✓ | | |
| [95] | | | ✓ | | ✓ | | |
| [66] | ✓ | | ✓ | | | | |
| [82] | ✓ | | ✓ | | ✓ | | |
| [100] | | ✓ | | | | | |

These values allow us to calculate the corresponding accuracy, precision, recall, and F-score. AUC (Area Under Curve) and ROC (Receiver Operating Characteristics), also called AUROC (Area Under Receiver Operating Characteristics), are calculated using TPR and FPR. The ROC curve is plotted with FPR on the x-axis against TPR on the y-axis. AUC is measured as the area under this curve. AUC values range as real values between 0 and 1, with values closest to 1 being good or correct classification while values closest to 0 being the worst with poor classification. EER, defined as Equal Error Rate, calculates the error that occurred in classification. Table 5 identifies the evaluation metrics used by the reviewed tasks.

## 6.2 Result Analysis

Several metrics and parameters have been developed to define the functional performance or, in simpler terms, the algorithms' efficiency on giving the desired output of classification of data from a given dataset. Among them and widely utilized and relied upon metrics are Accuracy (in percentage), Precision, Recall, and F1 values, among several others such as AUC and EER. Almost all major research related to Fake News detection utilizes one or more among the former set of four metrics (Accuracy, Precision, Recall, and F-Score).

Hence, we bring forth such metric evaluation summary of the most relevant and pivotal experiments conducted for fake news detection. Figure 17 demonstrates the results of tasks in terms of F1-scores. We observe that the overall performance stays between 80-95% for methods that use textual and visual features combined. The video classification task, which uses the annotation technique, still has a long way to go. In terms of accuracy (Figure 18), we observe that the range of results is between 70-100%, with an average score of 85%. The majority of fake news classification tasks have relied upon deep neural networks. With the changing time, we also notice an inclination towards forensic algorithms for the same. Trends determine that most of the existing approaches have preferred to use deep learning algorithms due to their efficiency, robust nature, feasibility, and accuracy. Most works have preferred to use more than one feature, i.e., using multi-modal data. Thus, depending on more options and the type of fake information posts can offer. The aim is to consider all parameters that form/alters a user's perception of a piece of information. Convolutional Neural Networks, with maximum usage in the reviewed articles, have displayed eminent classification performance by exploiting the implicit features. They hold the potential to provide better results in future implementations.

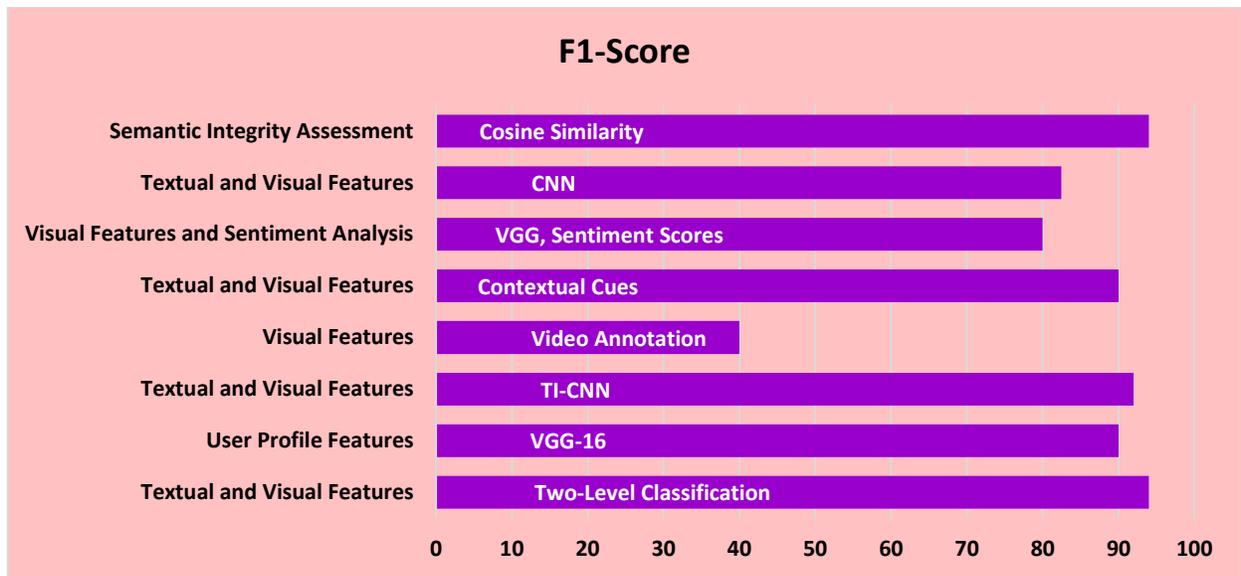

Figure17: Performance comparison based on F1 scores

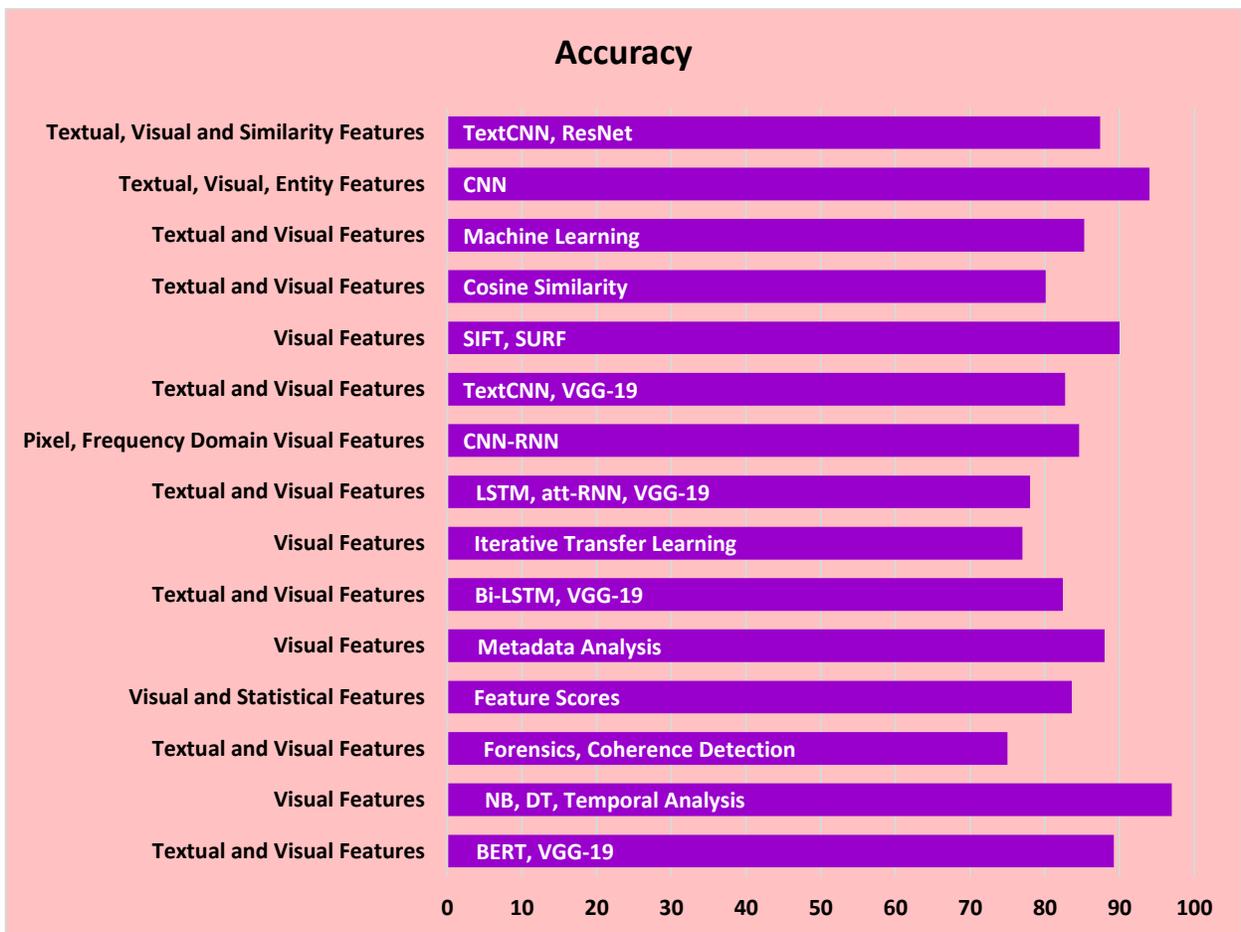

Figure 18: Performance comparison based on accuracy scores

## 7. Potential Research Directions

Research done in the past years is overwhelming yet, insufficient to cope with the amount of fake news pouring in. Each new happening or event in the world serves as a topic for fake news generation and propagation. In the present scenario, while a pandemic is going on, fake news reaches out to people more swiftly than authentic news is. No data modality is left behind in the race of spreading fake news. Text is not the only type of data of which one should be aware while intaking. People need to be more careful while digesting anything available on the internet because false information could greet us in any form, be it text, image, or video. So is the need for designing efficient and robust detection mechanisms. Analyzing the limitations and research gaps, this section highlights the potential directions where research can be proceeded into.

1. **Datasets:** From the above analysis, it is obvious that there is a shortage of large-scale multimodal datasets. Machine learning and Deep learning algorithms are data-driven. With the shortage of benchmark datasets, it becomes challenging to build detection mechanisms and compare various techniques' performances. Although there are a few text datasets, those with multimodal information are limited and of poor quality. With the advantage of web-scraping mechanisms and free APIs, it has become easier to collect data. To proceed in the direction of multimodality, the collection of large-scale multimodal datasets is promoted.

2. **Real-time Detection:** With the assistance of deep learning algorithms, real-time detection models can be built to use fact-checked articles on the web for training and generate predictions for unseen data. There is a wide opportunity for the development of real-time detectors and automated fact-checkers.

3. **Early Detection:** Fake news detectors are built by feeding past data to algorithms. A baseline comparison is made on previous data. These algorithms are built when the fake information has already spread into the world and affected many. Entrapped within the fake news web, the world requires early detection of false news as and when it appears online. Users can only be benefited from fake news detectors when they provide early detection to prevent the propagation of fake news to a large scale. Early detection would allow intervention and thus mitigation of fake news before it spreads to a larger audience.

4. **Ubiquitous Detection Model:** With many social networking platforms available, it is challenging to incorporate a fake news detection mechanism to separate platforms individually. Similar content makes rounds on multiple platforms because one user can have accounts on various networks. This creates a replica of data on different social networks. With the help of redundant data and manual annotations, classification becomes easy for deep neural networks. A cross-platform system is required that would detect fake content on multiple social platforms. Implementation of models that can train on manually annotated content on one platform and then identify fake news on other platforms is suggested.

5. **Data-oriented Detection:** As far as the previous research is considered, we have very few frameworks that provide credibility assessment to fake content types. Most techniques consider text only, while some allow visual verification. It is challenging for a single system to verify the contents of all data modalities. Such a system would be more beneficial for the general public to authenticate information.

6. **Feature-oriented Detection:** All existing approaches use a limited subset of features, either linguistics, visuals, hybrid, data-centered, sentiment scores, social context, network-based, user-based, or post-based features. These contributing factors of fake news identification could be used all together for dependable predictions.

7. **Integrity Assessment:** In multimodal approaches, existing works perform detection based on features from each type of data independently. In many fake news instances, the post contents are not semantically related. The text, image, or video for a given post could be expressing unrelated context. Few works focus on assessing the semantic integrity of the news. This helps detect false news where data modalities have not been manipulated but are unrelated to each other. Such integrity assessment tools shall help in identifying out-of-context news items.

8. **Embedded Fake News Detection:** Detection of fake embedded content has not been done yet. A large volume of fake news is spreading through such type of data. To cope up with the incoming fake news, this type of detection mechanism is required.

9. **Multilingual Detection:** Current approaches have focused on English language data in text and videos. Due to the spread of fake news through regional languages on the web, multi-lingual approaches should be considered to detect fake news from other languages in the form of text, videos, or embedded content.

10. **Data Manipulation Detection:** With the popularity of image and video forensics techniques, forgery detection in data has become easier. Various manipulation techniques like face-spoofing detection, deepfake identification, tampering detection, splicing, copy-move detection, object removal/addition detection, etc., should be considered and merged with fake news detection mechanisms. There is a need for merging the domains of fake news detection and data manipulation detection.

11. **Browser Plugin/Application Software:** The availability of fake news detector tools in the form of easy-to-use browser plugins, add-ons, software, and mobile applications will enhance their accessibility and serve detection on a user-basis.

## 8. Conclusion and Future Scope

Uncontrolled and unauthentic data being over-loaded on the web needs appropriate solutions for the complexities being generated and has become a hard nut to crack. Deep learning algorithms are proving efficient and providing effective solutions with remarkable results. These solutions are to be unearthed from unimaginable horizons and that too within a very precise and limited period as the flow of complexities has reached the verge of parallel solutions. There has been a rapid increase in luring solutions for multimodal fake news detection adopting numerous variant techniques. This survey allows us to conclude that deep learning architectures prove astonishingly capable of fake news detection. They have resulted in high accuracies under the text-domain. Recurrent Neural Networks, LSTMs, GRU, Bi-directional GRU have contributed significantly to text classification. When it comes to visual data, Convolutional Neural Networks form the bigger picture. Survey displays that over 40% of methodologies have incorporated CNNs and their combinations with RNNs or other DNNs in their detection frameworks and served brilliant results.

CNNs are taking the lead in computer vision, and allied domains and have become a prospective application for future FND tasks. Many researchers have identified fake images and videos and tampered regions in them, which we review as supportive tasks that can help classify fake news based on fake visuals. We motivate the readers to combine such tasks with FND modules to perform multimodal FND. By fusing modules performing such tasks on different modalities, optimized performances are assured.

There has been the unavailability of symbolic literature in this domain. The progress along the pathways of multimodal fake news detection has been slow. Researchers are unaware of the advancements so far reached. Existing literature is focused upon fake textual news and its detection mechanisms. This survey acknowledges this shortage and provides a broader and intact overview of multimodal fake news detection that incorporates image, video, audio, and their combinations with text. We engage the readers with a taxonomy of detection methods utilized in the referenced articles. We identify the methods with most applications and highlight their prospects. In this review, we have endeavored to cover almost all the significant works performed in relevant domains. We segregate articles based on modalities implied in them. We demonstrate the year-wise trend of work involved and also analyze their method-wise distribution. We neatly summarize all the notable work and highlight important techniques that may pose powerful algorithms in future fake news classification tasks. Further, we described the evaluation metrics adapted in research so far.

We see that accuracy, precision, recall, and F-scores were observed for most of the tasks, whereas some liked to evaluate their models' performances using AUC and ROC. A few other metrics utilized were EER, HTER, TPR, and FPR. Accuracy appears as the most adopted method. Further, owing to the scarcity of multimodal datasets, we regard the obstacles faced in fledged research in the form of not so optimum solutions. Hence, we provide collective information of all the good quality image, video, and multimodal datasets available and previously been used in various tasks. This provides a route for fellow researchers to efficiently choose amongst the few available datasets and perform future research. We also encourage them to build good quality multimodal fake news datasets that would render this domain positively.

Concerning future works, we promote a multimodal framework that would efficiently detect fake news in all forms that revolve around the internet. We suggest exploring the domain incorporating fake news in the form of videos. We also motivate researchers to engage in building versatile multimodal datasets for future use collecting information from websites, online social platforms, and the likes. We encourage the readers to dive deeper into machine learning and deep learning algorithms and fish out ultimate solutions to the problem domain. Our work helps in bridging the research gaps and serve as potential future opportunities to work upon. We conclude this review anticipating that interested researchers will benefit from the information provided and narrow down their interests to this domain to contribute to the society and research community.